\documentclass{article}
\usepackage{spconf,amsmath,graphicx}
\usepackage{booktabs}
\usepackage{siunitx}
\usepackage{url}
\usepackage{float}
\usepackage{enumitem}
\setlist{nosep, leftmargin=14pt}

\title{Detection and Classification of (Pre)Cancerous Cells in Pap Smears: An Ensemble Strategy for the RIVA Cervical Cytology Challenge}
\name{Lautaro Kogan$^{1}$ and María Victoria Ríos$^{1}$}
\address{$^{1}$Universidad de San Andrés, Buenos Aires, Argentina}

\begin{document}
%
\maketitle
\begin{abstract}
\small
Automated detection and classification of cervical cells in conventional Pap smear images can strengthen cervical cancer screening at scale by reducing manual workload, improving triage, and increasing consistency across readers. However, it is challenged by severe class imbalance and frequent nuclear overlap. We present our approach to the RIVA Cervical Cytology Challenge (ISBI 2026), which requires multi-class detection of eight Bethesda cell categories under these conditions. Using YOLOv11m as the base architecture, we systematically evaluate three strategies to improve detection performance: loss reweighting, data resampling and transfer learning. We build an ensemble by combining models trained under each strategy, promoting complementary detection behavior and combining them through Weighted Boxes Fusion (WBF). The ensemble achieves a mAP50-95 of 0.201 on the preliminary test set and 0.147 on the final test set, representing a 29\% improvement over the best individual model on the final test set and demonstrating the effectiveness of combining complementary imbalance mitigation strategies.
\end{abstract}

\section{Introduction}
\label{sec:intro}

Automated analysis of cytological images has the potential to substantially improve early detection 
of cervical precancer and cancer, particularly in low- and middle-income regions where manual 
screening remains the primary diagnostic modality. Conventional Pap smear interpretation is 
labor-intensive and subject to considerable inter-observer variability, which limits both throughput 
and diagnostic consistency. Recent advances in deep learning have achieved strong performance 
across a wide range of medical imaging tasks \cite{litjens2017survey}, yet their application to 
conventional cytology remains limited by the scarcity of large, well-annotated datasets.

The RIVA dataset addresses this gap as a large-scale collection of conventional Pap smear image 
patches annotated by up to four independent expert cytopathologists, enabling the study of inter-annotator variability and providing a realistic benchmark for automated nuclei detection and cell classification \cite{perez2025riva}. Building on this resource, the RIVA Cervical Cytology Challenge at ISBI 2026 focuses on automated nuclei detection and multi-class cell classification under severe class imbalance and frequent nuclear overlap \cite{riva-cervical-cytology-challenge-track-a-isbi-final-evaluation}. We present our approach to this 
challenge, systematically evaluating three training strategies — loss reweighting, transfer learning, and weighted sampling — within a unified YOLOv11m-based detection framework. Beyond optimizing challenge performance, we analyze the complementary generalization behavior of these strategies across challenge phases and propose to ensemble them via Weighted Boxes Fusion, achieving a 29\% improvement over the best individual model on the final test set.

\section{PROBLEM DEFINITION}
\label{sec:problem}

The RIVA dataset \cite{perez2025riva} is a collection of 959 high-resolution conventional Pap smear image fields (1024 × 1024 pixels, 40× magnification) from 115 patients, annotated by up to four expert cytopathologists following The Bethesda System (TBS). Cells are categorized into five (pre)cancerous types — ASCUS (atypical squamous cells of undetermined significance), LSIL (low-grade squamous intraepithelial lesion), ASCH (atypical squamous cells that cannot exclude high-grade lesion), HSIL (high-grade squamous intraepithelial lesion), and SCC (squamous cell carcinoma) — and three non-lesion categories — NILM (negative for intraepithelial lesion or 
malignancy), INFL (inflammatory cells), and ENDO (endocervical cells).

\begin{figure}[htb]
    \centering
    \includegraphics[width=1\linewidth]{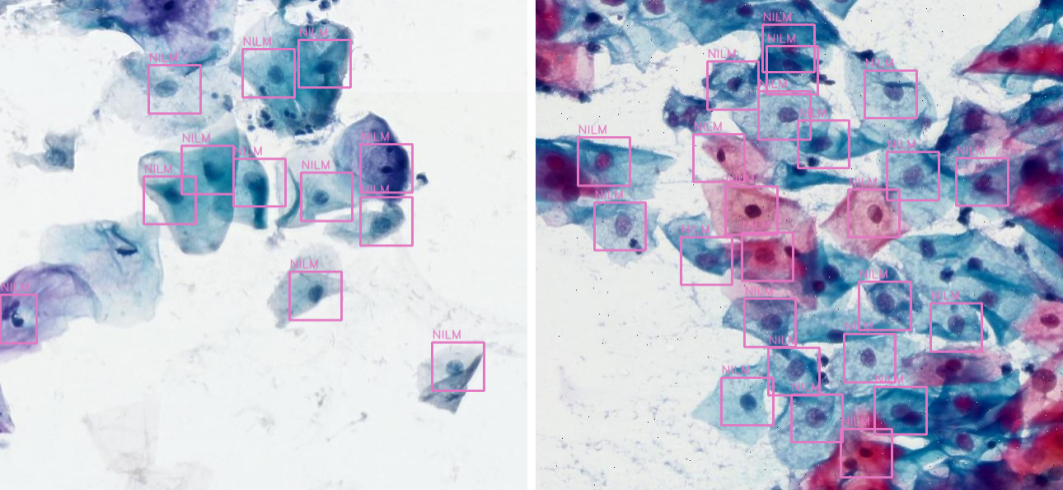}
    \caption{Examples of training images with ground-truth annotations.}
    \label{fig:examples}
\end{figure}

The dataset exhibits severe class imbalance, with malignant and high-grade lesions substantially underrepresented: 1,586 SCC, 1,835 HSIL, 416 ASCH, 3,048 LSIL, and 356 ASCUS instances, compared to 9,457 NILM, 8,190 INFL, and 1,270 ENDO cells. Additional challenges include frequent cell 
overlap, staining variability, and background artifacts. The RIVA Challenge defines two tasks: automated nuclei detection and multi-class classification into one of the eight Bethesda categories. 
Submissions are evaluated using mAP50-95, a metric stringent for overlapping nuclei. The dataset is split into training, validation, 
preliminary test, and final test sets, with annotations for the test sets withheld. A preliminary phase lasting over two months was followed by a one-week final phase upon release of the final test split. Representative training images are shown in Fig.~\ref{fig:examples}.

\section{METHODS}
\label{sec:methods}

\subsection{Training and Inference Protocol}
\label{ssec:protocol}
Model development and fine-tuning were carried out exclusively during the preliminary phase, using the training, validation, and preliminary test splits to guide architectural and hyperparameter decisions. No further training was performed after the preliminary phase ended. 

\subsection{Data pre-processing}
\label{ssec:data-pre-processing}
Preprocessing was applied to training data only, including automatic orientation correction and data augmentations consisting of horizontal and vertical flips and additive noise, expanding each original image into three augmented samples. Despite these augmentations, the training data 
exhibits severe class imbalance, motivating the mitigation strategies described below.

\subsection{Baseline YOLO configuration and class imbalance mitigation strategies}
\label{ssec:baseline-yolo-configuration}

We adopted YOLOv11m as the base detection architecture for all experiments, implemented using the Ultralytics framework \cite{ultralytics_yolov8}, due to its favorable trade-off between 
accuracy and computational efficiency in dense detection scenarios \cite{khanam2024yolov11}. We trained three models following different optimization strategies to promote complementary behavior, enabling their combination within an ensemble weighted by 
final test set mAP50-95.

\subsubsection{Model 1: Loss reweighting} 
\label{sssec:model-1}

The first strategy addresses class imbalance directly at the loss level, following weighted cross-entropy formulations commonly used to mitigate class imbalance in object detection \cite{phan2020weighted}. Given the highly skewed class distribution observed in the training data, class-specific weights are applied to the classification component of the loss function to emphasize rare lesion categories during optimization. Let $p_k$ denote the relative frequency of the class $k$ in the training set. Class weights are computed as: 
\begin{equation}
    w_k = -\log(p_k).
\end{equation}

These weights are incorporated into the binary cross-entropy–based classification loss used by YOLO through a weighted formulation, increasing the contribution of underrepresented classes while reducing the risk of excessive gradient amplification. The logarithmic scaling provides a smoother rebalancing effect, improves numerical stability, and helps mitigate overfitting to majority classes.

As reported in Table 1, this approach achieved a mAP50-95 of 0.160 on the preliminary test set. As reported in Table 2, it achieved a mAP50-95 of 0.108 on the final test set.

\subsubsection{Model 2: Transfer learning from SIPaKMeD}
\label{sssec:model-2}

The second model employs transfer learning using the SIPaK\-MeD dataset, a public resource containing 966 image patches and 4,049 isolated cervical cell images \cite{plissiti2018sipakmed}. The YOLOv11m model was first pre-trained on SIPaKMeD, then fine-tuned on the RIVA dataset with the 10 backbone layers frozen. This allows the model to leverage low-level and morphological features inherent to cervical cytology, providing a domain-relevant initialization. It also exhibited the smallest train-test performance gap during the preliminary phase, suggesting improved generalization compared to the alternative configurations. As shown in Table 1, this model achieved a slightly lower mAP50-95 than Model 1 during the preliminary phase. On the final test set, it obtained a mAP50-95 of 0.106, as reported in Table 2.

\subsubsection{Model 3: Weighted data loader }
\label{sssec:model-3}

The third model addresses class imbalance using weighted sampling at the dataloader level, following resampling approaches commonly applied in imbalanced object detection \cite{yasin2024weighted}. Instead of modifying the loss function, the sampling probability of each training image is adjusted based on the class distribution of the objects it contains. For each class $k$ with count $c_k$, we compute class weights as
\begin{equation}
    w_k = \sqrt{\frac{\sum_i c_i}{c_k}},
\end{equation}

\noindent where $\sum c_i$ is the total number of objects across all classes. Each image receives a sampling weight calculated by aggregating the weights of all classes it contains. In our implementation, the aggregation is performed using the arithmetic mean. Images are then sampled with probability proportional to these weights, increasing exposure of underrepresented lesion classes while maintaining dataset diversity. The square-root formulation smooths extreme weights, reducing the risk of overfitting to minority classes. This reweighting strategy is applied only during training.

As shown in Table 1, this model achieved the lowest mAP50-95 among the individual models on the preliminary test set (0.127) and exhibited a pronounced train-test gap, indicative of overfitting. Nonetheless, it achieved the highest mAP50-95 on the final test set (0.114), as reported in Table 2. Additionally, it demonstrated high recall for critical lesions on the validation set, achieving 49\% recall for SCC and highlighting its ability to detect high-risk lesions. 

\subsubsection{Ensemble strategy and post-processing}
\label{sssec:ensemble}

To improve detection robustness and leverage complementary strengths across models, we combine the predictions of the three independently trained YOLO models using Weighted Boxes Fusion (WBF) \cite{solovyev2021wbf} \cite{xu2019ensemble}. For each test image, bounding box predictions, confidence scores, and class labels from all models are merged through this fusion process. Each model is assigned a weight equal to its mAP50-95 on the final test set, specifically 0.108, 0.106, and 0.114 for Models 1, 2, and 3 respectively, such that models with stronger standalone performance exert proportionally greater influence on the fused output. Bounding boxes are fused when their IoU exceeds or equals 0.7. This relatively strict threshold is specifically adopted to handle the frequent nuclear overlap inherent to the RIVA dataset, preventing distinct adjacent cells from being merged while still consolidating redundant model predictions to improve spatial precision for the mAP50-95 metric. Low-confidence predictions are filtered using a minimum confidence threshold of 0.001 to maintain recall for underrepresented lesion classes.

\section{Results}
\label{sec:results}

Performance is evaluated across two complementary dimensions: challenge-protocol mAP50-95, which quantifies detection accuracy under the official metric, and per-class Precision--Recall analysis, which provides insight into model behavior for clinically critical minority classes beyond what the aggregate metric captures.

For the challenge metric, mAP50-95 is computed using a confidence threshold of 0.001, following the official evaluation protocol. Under this setting, the ensemble model achieves a mAP50-95 of 0.201 on the preliminary test set, consistently outperforming all individual models (Table~\ref{tab:map_results_prelim}).

\begin{table}[htb]
\centering
\caption{Model performance on the preliminary test set measured by mAP50-95.}
\label{tab:map_results_prelim}
\resizebox{\columnwidth}{!}{%
\begin{tabular}{l c c}
\toprule
\textbf{Model} & \textbf{mAP50-95 (Test)} & \textbf{Gap w/ Train} \\
\midrule
1. Loss Reweighting    & 0.160 & -0.05 \\
2. Transfer Learning   & 0.152 & -0.01 \\
3. Weighted Dataloader & 0.127 & -0.06 \\
\textbf{Ensemble}      & 0.201 & -- \\
\bottomrule
\end{tabular}%
}
\end{table}

On the final test set, the ensemble also outperforms all individual models, achieving 0.147, representing a 29\% improvement over the best individual model (Table~\ref{tab:map_results_final}).

\begin{table}[H]
\centering
\caption{Model performance on the final test set measured by mAP50-95.}
\label{tab:map_results_final}
\resizebox{\columnwidth}{!}{%
\begin{tabular}{l c c}
\toprule
\textbf{Model} & \textbf{mAP50-95 (Test)} & \textbf{Weight on Ensemble} \\
\midrule
1. Loss Reweighting    & 0.108 & 0.108 \\
2. Transfer Learning   & 0.106 & 0.106 \\
3. Weighted Dataloader & 0.114 & 0.114 \\
\textbf{Ensemble}      & 0.147 & -- \\
\bottomrule
\end{tabular}%
}
\end{table}

Notably, the ranking of individual models shifts between the preliminary and final test sets: Model~3 (Weighted Dataloader), despite exhibiting the lowest preliminary test mAP50-95 and a pronounced train-test gap suggestive of overfitting, achieves the highest mAP50-95 among individual models on the final test set (0.114). This reversal highlights the risk of relying solely on preliminary phase results to draw conclusions about generalization. Model~2 (Transfer Learning), by contrast, exhibited the smallest train-test gap throughout the preliminary phase, consistent with its domain-relevant initialization, though this did not translate into the highest final test performance. These behavioral differences across phases motivated the ensemble design, where models were weighted by final test performance to reflect their true generalization rather than their preliminary rankings.

In a second stage, the Precision--Recall (PR) curve is analyzed using the validation set, as ground-truth annotations for the test sets are not publicly available. This analysis complements the mAP metric by revealing per-class detection behavior, particularly for underrepresented lesion categories where aggregate metrics can be misleading. Precision and recall are computed by matching predictions to ground-truth boxes at $IoU \ge 0.5$, sweeping across all confidence levels to trace the full curve. The area under each curve is reported as the Average Precision (AP) for each class.

\begin{figure}[htb]
    \centering
    \includegraphics[width=1\linewidth]{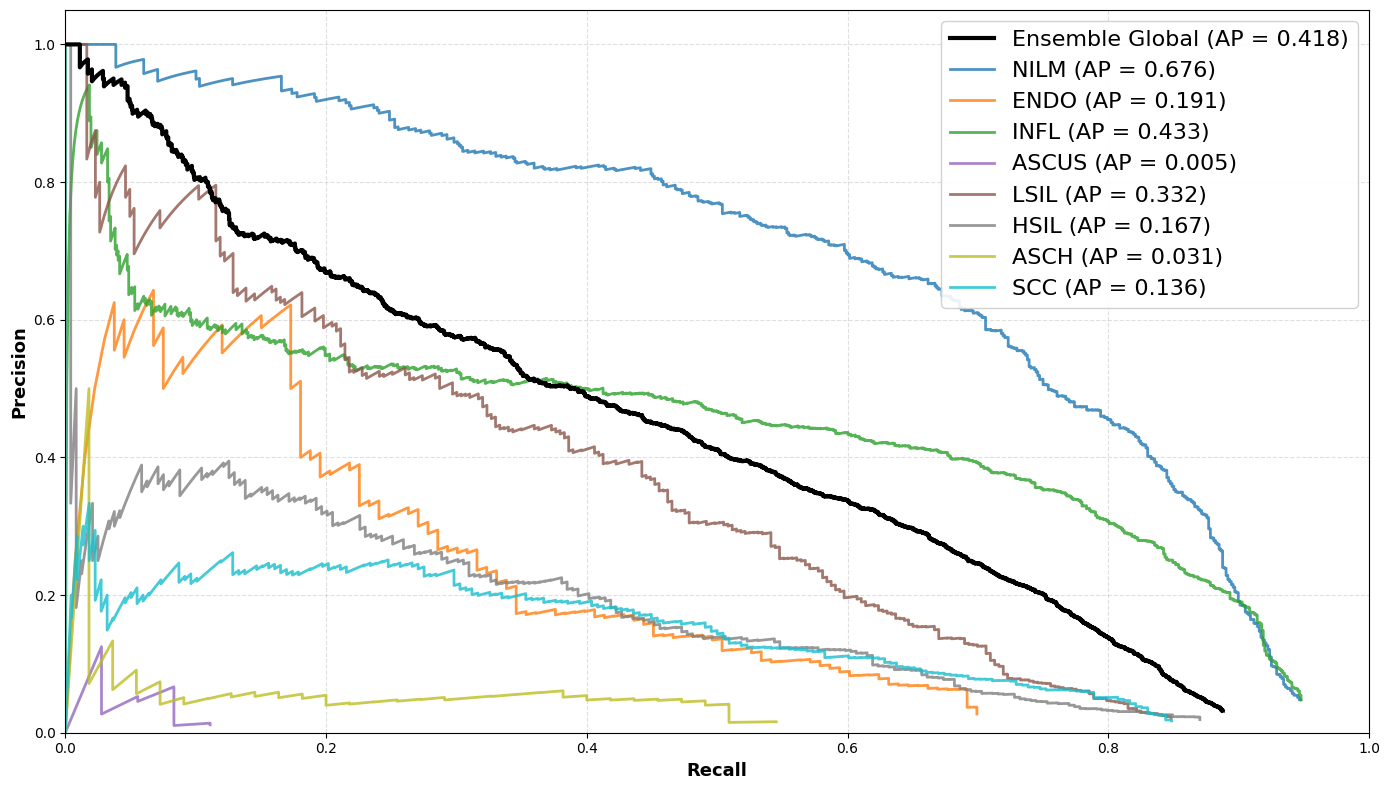}
    \caption{Precision--Recall curve of the ensemble model evaluated on the validation set (IoU=0.5).}
    \label{fig:pr_curve}
\end{figure}

Fig.~\ref{fig:pr_curve} presents the PR curve of the ensemble, achieving a global Average Precision (AP) of 0.418. Precision remains high at higher confidence levels, critical for computer-assisted screening, and gradually declines as recall increases, effectively recovering difficult detections in minority classes. Per-class performance shows strong correlation with class imbalance: NILM (AP = 0.676, n = 748) and INFL (AP = 0.433, n = 863) achieve acceptable results, while rare classes ASCUS (AP = 0.005, n = 36) and ASCH (AP = 0.031, n = 55) suffer from extreme imbalance and morphological ambiguity. At the F1-optimal threshold (0.1), ASCH achieves 38.8\% recall with 6\% precision, functioning as a sensitivity-first pre-screening tool requiring expert validation. 

\section{Conclusion}
\label{sec:conclusion}

This work presents an approach to automated detection and classification of cervical precancerous cells under severe class imbalance, achieving a final test set mAP50-95 of 0.147 through ensemble strategies. Beyond optimizing challenge performance, a central contribution of this work is the analysis of complementary generalization behavior across three training strategies, showing that preliminary phase rankings are not reliable predictors of final test generalization and that different strategies offer distinct clinical trade-offs between minority class sensitivity and overall detection stability. The ensemble via WBF leveraged these complementary strengths, increasing final test set performance by 29\% over the best individual model.

However, significant challenges remain, particularly for rare but clinically critical classes (ASCUS: AP = 0.006, ASCH: AP = 0.045). The low WBF threshold (0.001) necessary to maximize recall for minority classes introduces substantial visual noise, requiring careful threshold 
adjustment for clinical deployment. Future work should explore advanced minority class augmentation, class-specific confidence thresholds, and two-stage detection-classification pipelines to improve performance on rare lesion categories.

\section{Compliance with Ethical Standards}
This research study was conducted retrospectively using human subject data made available in open access by Perez Bianchi et al. \cite{perez2025riva}. Ethical approval was not required as confirmed by the license attached with the open access data.

\section{Acknowledgments}
No funding was received for conducting this study. The authors have no relevant 
financial or non-financial interests to disclose.

\bibliographystyle{IEEEbib}
\bibliography{strings,refs}

\end{document}